\documentclass{article}

\usepackage{graphicx}        
\usepackage{multicol}        
\usepackage[bottom]{footmisc}

\usepackage{ amssymb }

\newcommand{\myvcenter}[1]{\ensuremath{\vcenter{\hbox{#1}}}}

\makeindex             

\begin{document}

\title{Bayesian Inference in \\Cumulative Distribution Fields}
\author{Ricardo Silva\\University College London, UK \\ \tt{ricardo@stats.ucl.ac.uk}}
%
%
\maketitle

\abstract{One approach for constructing copula functions is by
multiplication. Given that products of cumulative distribution
functions (CDFs) are also CDFs, an adjustment to this
multiplication will result in a copula model, as discussed by
Liebscher (J Mult Analysis, 2008). Parameterizing models via
products of CDFs has some advantages, both from the copula perspective
(e.g., it is well-defined for any dimensionality) and from general
multivariate analysis (e.g., it provides models where small dimensional
marginal distributions can be easily read-off from the parameters).
Independently, Huang and Frey (J Mach Learn Res, 2011) showed the
connection between certain sparse graphical models and products of
CDFs, as well as message-passing (dynamic programming) schemes for
computing the likelihood function of such models. Such schemes allows
models to be estimated with likelihood-based methods. We discuss and
demonstrate MCMC approaches for estimating such models in a Bayesian
context, their application in copula modeling, and how
message-passing can be strongly simplified. Importantly, our view of
message-passing opens up possibilities to scaling up such methods,
given that even dynamic programming is not a scalable solution for calculating
likelihood functions in many models.}

\section{Introduction}
\label{sec:1}

Copula functions are cumulative distribution functions (CDFs) in the
unit cube $[0, 1]^p$ with uniform marginals. Copulas allow for the
construction of multivariate distributions with arbitrary marginals --
a result directly related to the fact that $F(X)$ is uniformly
distributed in $[0, 1]$, if $X$ is a continuous random variable with
CDF $F(\cdot)$. The space of models includes semiparametric models,
where infinite-dimensional objects are used to represent the
univariate marginals of the joint distribution, while a convenient
parametric family provides a way to represent the dependence
structure. Copulas also facilitate the study of measures of dependence
that are invariant with respect to large classes of transformations of
the variables, and the design of joint distributions where the degree
of dependence among variables changes at extreme values of the sample
space. For a more detailed overview of copulas and its uses, please
refer to \cite{joe:97,nelsen:07,elidan:14}.

A multivariate copula can in theory be derived from any joint
distribution with continuous marginals: if $F(X_1, \dots, X_p)$ is a
joint CDF and $F_i(\cdot)$ is the respective marginal CDF of $X_i$,
then $F(F_1^{-1}(\cdot), \dots, F_p^{-1}(\cdot))$ is a copula. A
well-known result from copula theory, Sklar's theorem \cite{nelsen:07},
provides the general relationship. In practice, this requires being
able to compute $F_i^{-1}(\cdot)$, which in many cases is not a
tractable problem. Specialized constructions exist, particularly
for recipes which use small dimensional copulas as building blocks.
See \cite{bedford:02,kirshner:07} for examples. 

In this paper, we provide algorithms for performing Bayesian inference
using the product of copulas framework of Liebscher
\cite{liebscher:08}. Constructing copulas by multiplying functions of
small dimensional copulas is a conceptually simple construction, and
does not require the definition of a hierarchy among observed
variables as in \cite{bedford:02} nor restricts the possible structure
of the multiplication operation, as done by \cite{kirshner:07} for the
space of copula densities that must obey the combinatorial structure
of a tree. Our contribution is computational: since a product of
copulas is also a CDF, we need to be able to calculate the likelihood
function if Bayesian inference is to take
place\footnote{Pseudo-marginal appproaches \cite{andrieu:09}, which
use estimates of the likelihood function, are discussed briefly in the
last Section.}. The structure of our contribution is as follows: i. we
simplify the results of \cite{huang:10a}, by reducing them to standard message passing
algorithms as found in the literature of graphical models
\cite{cowell:99} (Section \ref{sec:message-passing}); ii. for intractable likelihood problems, 
an alternative latent variable representation for the likelihood
function is introduced, following in spirit the approach of
\cite{walker:11} for solving doubly-intractable Bayesian inference
problems by auxiliary variable sampling (Section
\ref{sec:auxiliary-variable}).

We start with Section \ref{sec:cdf-fields}, where we discuss with some
more detail the product of copulas representation. Some illustrative
experiments are described in Section \ref{sec:experiments}. We
emphasize that our focus in this short paper is computational, and
we will not provide detailed applications of such models. Some
applications can be found in \cite{huang:11}.

\section{Cumulative Distribution Fields}
\label{sec:cdf-fields}

Consider a set of random variables $\{U_1, \dots, U_p\}$, each having
a marginal density in $[0, 1]$. Realizations of this distribution are represented
as $\{u_1, \dots, u_p\}$. Consider the problem of defining a
copula function for this set. The product of two or more CDFs is a
CDF, but the product of two or more copulas is in general not a copula
-- marginals are not necessarily uniform after multiplication. In
\cite{liebscher:08}, different constructions based on products of
copulas are defined so that the final result is also a copula. In
particular, for the rest of this paper we will adopt the construction
\begin{equation}
\label{eq:basic-cdf}
C(u_1, \dots, u_p) \equiv \prod_{j = 1}^K C_j(u_1^{a_{1j}}, \dots, u_p^{a_{pj}})
\end{equation}
\noindent where $a_{i1} + \dots + a_{iK} = 1$, $a_{ij} \ge 0$ for all $1 \leq i \leq p$,
$1 \leq j \leq K$, with each $C_j(\cdot, \dots, \cdot)$ being a copula function.

Independently, Huang and Frey \cite{huang:08, huang:11} derived a
product of CDFs model from the point of view of graphical models,
where independence constraints arise due to the absence of some
arguments in the factors (corresponding in (\ref{eq:basic-cdf}) to
setting some exponents $a_{ij}$ to zero). Independence constraints from
such models include those arising from models of marginal independence
\cite{drton:03,drton:08}. 
\vspace{0.1in}

\noindent{\bf Example 1} We first adopt the graphical notation of \cite{drton:03} to describe
the factor structure of the cumulative distribution network (CDN)
models of Huang and Frey, where a bi-directed edge $U_m \leftrightarrow
U_n$ is included if $U_m$ and $U_n$ appear together as arguments to any
factor in the joint CDF product representation. For instance,
for the model $C(u_1, u_2, u_3) \equiv C_1(u_1, u_2^{1/2})C_2(u_2^{1/2},u_3)$
we have the corresponding network
\[
U_1 \leftrightarrow U_2 \leftrightarrow U_3
\]
First, we can verify this is a copula function by calculating the
univariate marginals.  Marginalization is a computationally trivial
operation in CDFs: since $C(u_1, u_2, u_3)$ means the probability
$P(U_1 \leq u_1, U_2, \leq u_2, U_3 \leq u_3)$, one can find the
marginal CDF of $U_1$ by evaluating $C(u_1, \infty, \infty)$. One can
then verify that $P(U_i \leq u_i) = u_i$, $i = \{1, 2, 3\}$, which is
the CDF of an uniform random variable given that $u_i \in [0, 1]$.
One can also verify that $U_1$ and $U_3$ are marginally independent
(by evaluating $C(u_1, \infty, u_3)$ and checking it factorizes), but
that in general $U_1$ and $U_3$ are {\it not} conditionally independent given $U_2$.
$\blacksquare$
\vspace{0.1in}

See \cite{drton:03, drton:08, huang:11} for an in-depth discussion of
the independence properties of such models, and \cite{liebscher:08}
for a discussion of the copula dependence properties. Such copula models
can also be defined conditionally. For a (non-Gaussian) multiple
regression model of outcome vector $\mathbf Y$ on covariate vector
$\mathbf X$, a possible parameterization is to define the density of
$p(y_i\ |\ \mathbf x)$ and the joint copula $C(U_1, \dots, U_p)$ where $U_i \equiv P(Y_i
\leq y_i\ |\ \mathbf x)$. Copula parameters can also be functions
of $\mathbf X$.

Bayesian inference can be performed to jointly infer the posterior
distribution of marginal and copula parameters for a given
dataset. For simplicity of exposition, from now on we will assume our
data is continuous and follows univariate marginal distributions in
the unit cube. We then proceed to infer posteriors over copula
parameters only\footnote{In practice, this could be achieved by
fitting marginal models $\hat{F}_i(\cdot)$ separately, and
transforming the data using plug-in estimates as if they were the true
marginals. This framework is not uncommon in frequentist estimation
of copulas for continuous data, popularized as ``inference function for margins'', IFM
\cite{joe:97}.}. We will also assume that for regression models the copula
parameters do not depend on the covariate vector $\mathbf x$. The terms
``cumulative distribution network'' (CDN) and ``cumulative distribution fields''
will be used interchangeably, with the former emphasizing the independence
properties that arise from the factorization of the CDF.

\section{A Dynamic Programming Approach for Aiding MCMC}
\label{sec:message-passing}

Given the parameter vector $\theta$ of a copula function and data
$\mathcal D \equiv \{\mathbf{U}^{(1)}, \dots, \mathbf{U}^{(N)}\}$, we
will describe Metropolis-Hastings approaches for generating samples
from the posterior distribution $p(\theta\ |\ \mathcal D)$. The
immediate difficulty here is calculating the likelihood function,
since (\ref{eq:basic-cdf}) is a CDF function. Without further information
about the structure of a CDF, the computation of the corresponding probability
density function (PDF) has a cost that is exponential in the
dimensionality $p$ of the problem. The idea of a CDN is to be able to provide
a computationally efficient way of performing this operation if the factorization
of the CDF has a special structure. 
\vspace{0.1in}

\noindent{\bf Example 2} Consider a ``chain-structured'' copula
function given by $C(u_1, \dots, u_p) \equiv C_1(u_1,
u_2^{1/2})C_2(u_2^{1/2}, u_3^{1/2}) \dots C_{p - 1}(u_{p - 1}^{1/2},
u_p)$. We can obtain the density function $c(u_1, \dots, u_p)$ as
\[
\begin{array}{rcl}
c(u_1, \dots, u_p) &=& 
     \displaystyle
     \left[\frac{\partial^2 C_1(u_1, u_2^{1/2})}{\partial u_1\partial u_2}\right]
     \left[\frac{\partial^{p - 2} C_2(u_2^{1/2}, u_3^{1/2}) \dots C_{p - 1}(u_{p - 1}^{1/2}, u_p)}{\partial u_3 \dots \partial u_p}\right] + \\
      &&
     \displaystyle
     \left[\frac{\partial C_1(u_1, u_2^{1/2})}{\partial u_1}\right]
     \left[\frac{\partial^{p - 1} C_2(u_2^{1/2}, u_3^{1/2}) \dots C_{p - 1}(u_{p - 1}^{1/2}, u_p)}{\partial u_2 \dots \partial u_p}\right]\\
      & \equiv &
    \displaystyle
    \frac{\partial^2 C_1(u_1, u_2^{1/2})}{\partial u_1\partial u_2} \times m_{2 \rightarrow 1}(u_2) +
    \frac{\partial C_1(u_1, u_2^{1/2})}{\partial u_1} \times m_{2 \rightarrow 1}(\bar{u}_2)
\end{array}
\]

\noindent Here, $m_{2 \rightarrow 1} \equiv [m_{2 \rightarrow 1}(u_2)\
m_{2 \rightarrow 1}(\bar{u}_2)]^{\mathsf T}$ is a two-dimensional
vector corresponding to the factors in the above derivation, known in
the graphical modeling literature as a {\it message} \cite{cowell:99}.
Due to the chain structure of the factorization, computing this vector
is a recursive procedure. For instance,
\[
\begin{array}{rcl}
m_{2 \rightarrow 1}(u_2) &=& 
     \displaystyle
     \left[\frac{\partial C_2(u_2^{1/2}, u_3^{1/2})}{\partial u_3}\right]
     \left[\frac{\partial^{p - 3} C_3(u_3^{1/2}, u_4^{1/2}) \dots C_{p - 1}(u_{p - 1}^{1/2}, u_p)}{\partial u_4 \dots \partial u_p}\right] + \\
      &&
     \displaystyle
     \left[C_2(u_2^{1/2}, u_3^{1/2})\right]
     \left[\frac{\partial^{p - 2} C_3(u_3^{1/2}, u_4^{1/2}) \dots C_{p - 1}(u_{p - 1}^{1/2}, u_p)}{\partial u_3 \dots \partial u_p}\right] \\
      & \equiv &
    \displaystyle
    \frac{\partial C_2(u_2^{1/2}, u_3^{1/2})}{\partial u_3} \times m_{3 \rightarrow 2}(u_3) +
    C_2(u_2^{1/2}, u_3^{1/2}) \times m_{3 \rightarrow 2}(\bar{u}_3)
\end{array}
\]
\noindent implying that computing the two-dimensional vector $m_{2 \rightarrow 1}$ corresponds to a summation of two terms, once we have
pre-computed $m_{3 \rightarrow 2}$. This recurrence relationship corresponds to a $\mathcal O(p)$ dynamic programming algorithm. $\blacksquare$
\vspace{0.1in}

The idea illustrated by the above example generalizes to trees and
junction trees. The generalization is implemented as a message passing
algorithm by \cite{huang:08, huang:10a} named the {\it
derivative-sum-product} algorithm.  Although \cite{huang:08}
represents CDNs using {\it factor graphs} \cite{kschis:01}, neither
the usual independence model associate with factor graphs holds in
this case (instead the model is equivalent to other already existing
notations, as the bi-directed graphs used in \cite{drton:03}), nor the
derivative-sum-product algorithm corresponds to the standard
sum-product algorithms used to perform marginalization operations in factor graph models.
Hence, as stated, the derivative-sum-product algorithm requires new
software, and new ways of understanding approximations when the graph
corresponding to the factorization has a high treewidth, making
junction tree inference intractable \cite{cowell:99}. In particular,
in the latter case Bayesian inference is doubly-intractable (following
the terminology introduced by \cite{murray:06}) since the likelihood
function cannot be computed.

Neither the task of writing new software nor deriving new
approximations are easy, with the full junction tree algorithm of
\cite{huang:10a} being considerably complex\footnote{Please notice that
\cite{huang:10a} also presents a way of calculating the gradient of
the likelihood function within the message passing algorithm, and as
such has also its own advantages for tasks such as maximum likelihood
estimation or gradient-based sampling. We do not cover gradient
computation in this paper.}. In the rest of this Section, we show a
simple recipe on how to reduce the problem of calculating the PDF
of a CDN to the standard sum-product problem.

Let (\ref{eq:basic-cdf}) be our model. Let $\mathbf z$ be a
$p$-dimensional vector of integers, each $z_i \in \{1, 2, ..., K\}$.
Let $\mathcal Z$ be the $p^K$ space of all possible assigments of
$\mathbf z$. Finally, let $I(\cdot)$ be the indicator function, where
$I(x) = 1$ if $x$ is a true statement, and zero otherwise.

The chain rule states that
\begin{equation}
\label{eq:pre_joint_z}
\frac{\partial^p C(u_1, \dots, u_p)}{\partial u_1 \dots \partial u_p} = 
           \sum_{\mathbf{z} \in \mathcal Z}
                 \prod_{j = 1}^K \phi_j(\mathbf u, \mathbf z)
\end{equation}

\noindent where 
\[
\phi_j(\mathbf u, \mathbf z) \equiv
\frac{\partial^{\sum_i^p I(z_i = j)} C_j(u_1^{a_{1j}}, \dots, u_p^{a_{pj}})}
     {\prod_{i\ s.t. \ z_i = j} \partial u_i}
\]

\noindent To clarify, the set $i\ s.t.\ z_i = j$ are the indices of the
set of variables $\mathbf z$ which are assigned the value of $j$
within the particular term in the summation.

From this, we interpret the function
\begin{equation}
\label{eq:joint_z}
p_c(\mathbf u, \mathbf z) \equiv \prod_{j = 1}^K \phi_j(\mathbf u, \mathbf z)
\end{equation}

\noindent as a joint density/mass function over the space $[0, 1]^p
\times \{1, 2, \dots, K\}^p$ for a set of random variables $\mathbf U \cup \mathbf Z$. 
This interpretation is warranted by the
fact that $p_c(\cdot)$ is non-negative and integrates to 1. For the
structured case, where only a subset of $\{U_1, \dots, U_p\}$ are arguments
to any particular copula factor $C_j(\cdot)$, the corresponding sampling space
of $z_i$ is $\mathcal Z_i \subseteq \{1, 2, \dots, K\}$, 
the indices of the factors which are functions of $U_i$. This follows from the fact
that for a variable $y$ unrelated to $\mathbf x$ we have $\partial f(\mathbf x) / \partial y = 0$, and as such for $z_i = j$ we have 
$\phi_j(\mathbf u, \mathbf z) = p_c(\mathbf u, \mathbf z) = 0$ if $C_j(\cdot)$ does not vary with $u_i$.
From this, we also generalize the definition of $\mathcal Z$ to $\mathcal Z_1 \times \dots \times \mathcal Z_p$.

\begin{figure}[t]
\begin{center}
\begin{tabular}{cc}
\myvcenter{\includegraphics[scale=.35]{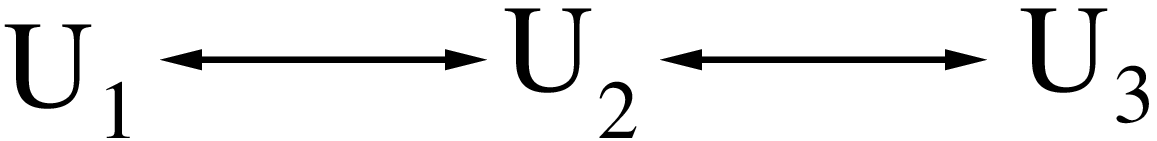}} &
\myvcenter{\includegraphics[scale=.35]{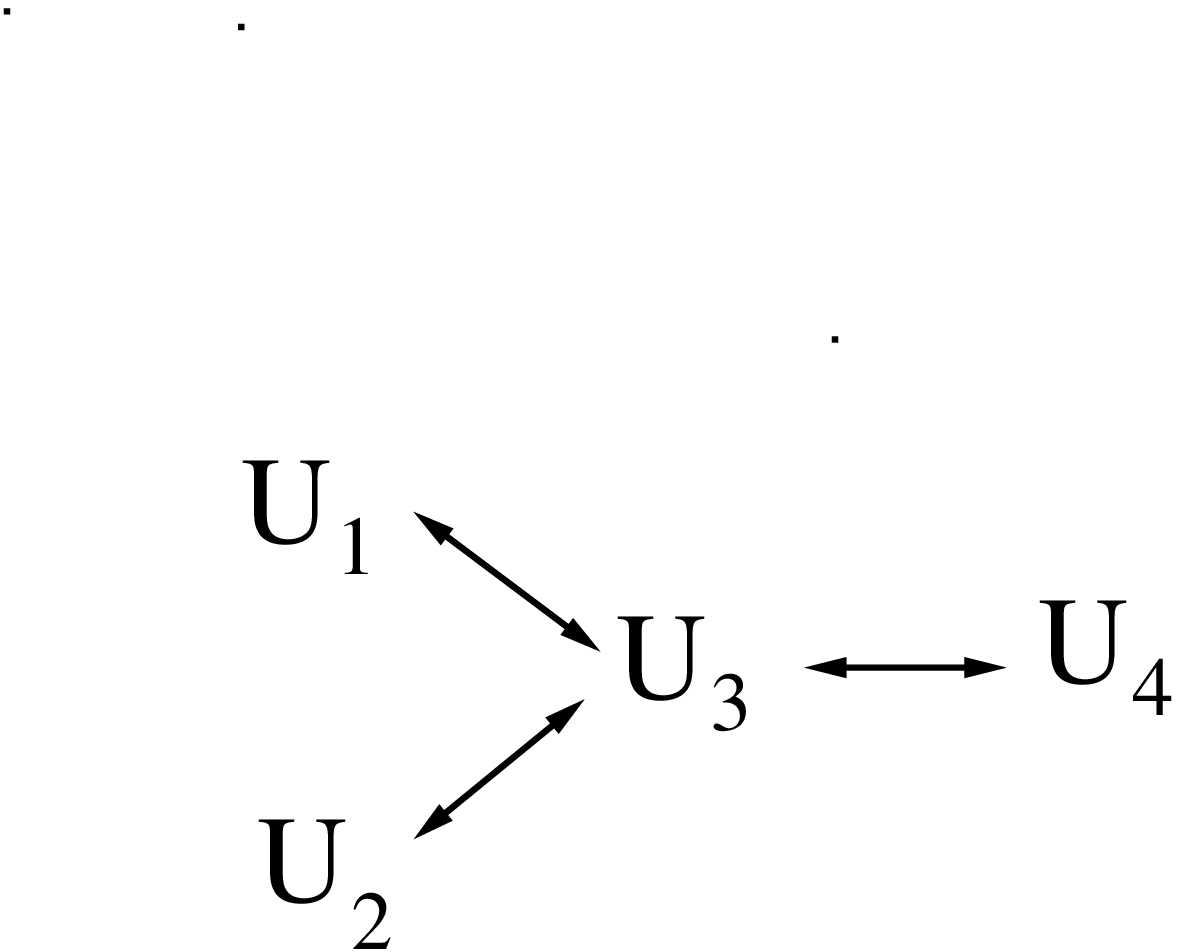}} \\
(a) & (b) \\
\myvcenter{\includegraphics[scale=.35]{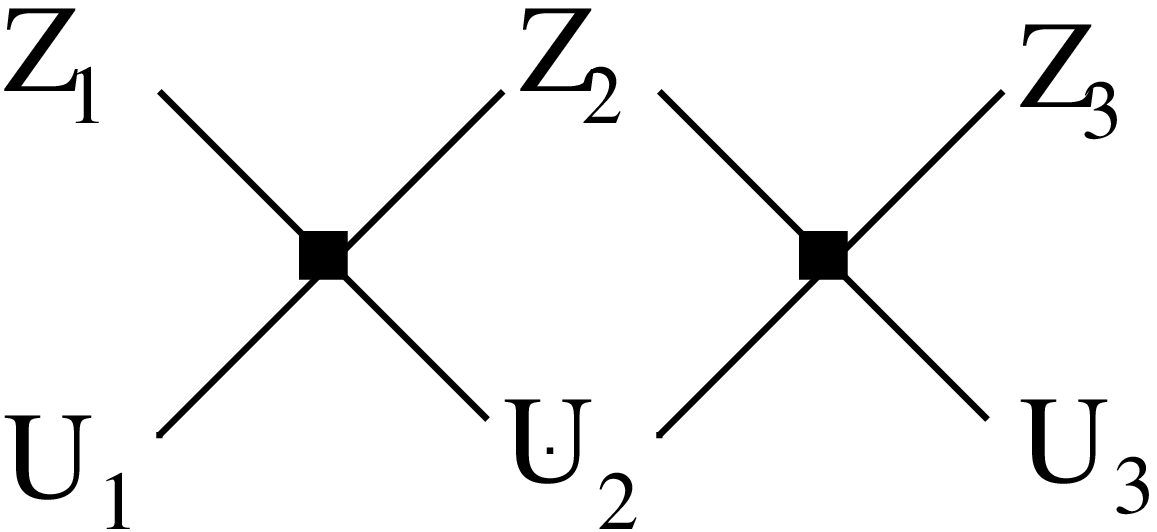}} &
\myvcenter{\includegraphics[scale=.35]{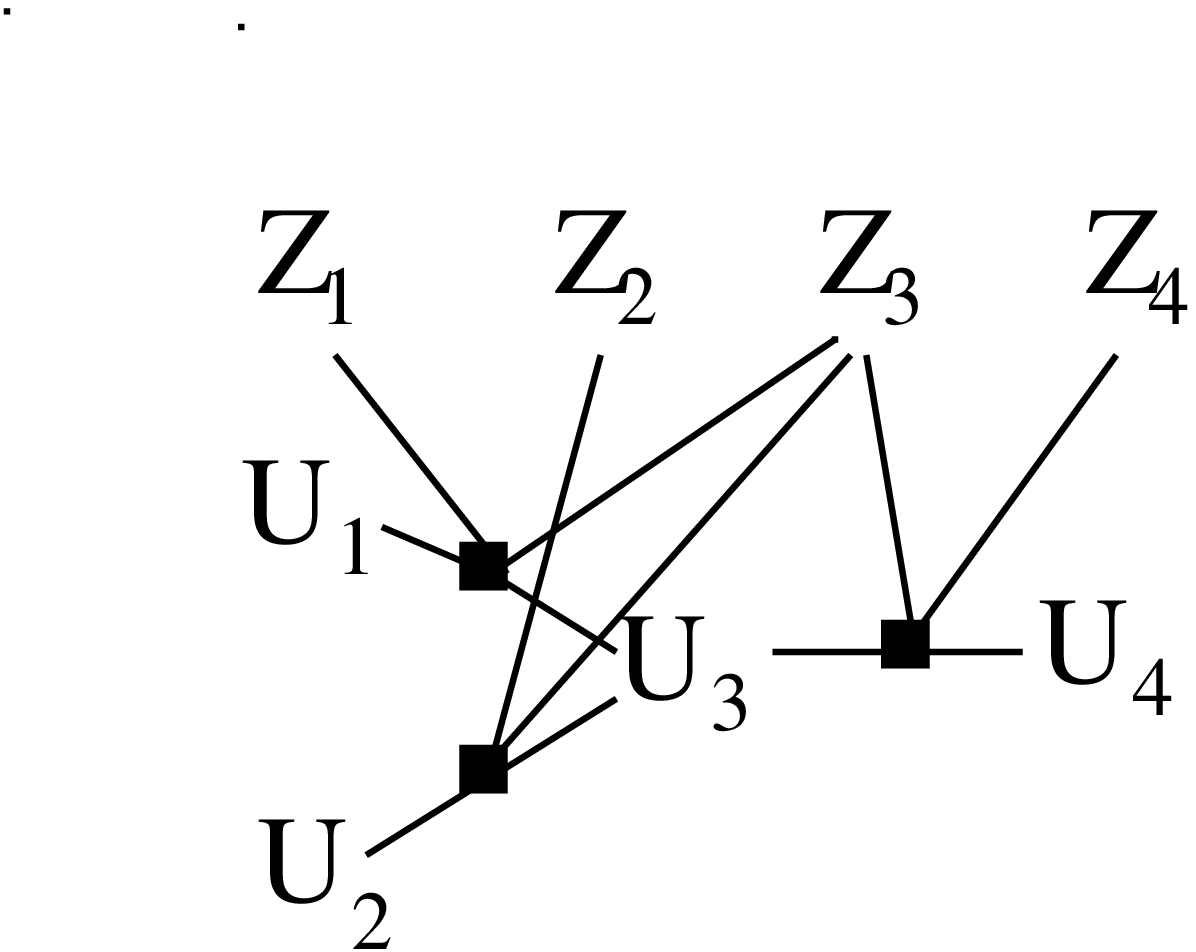}} \\
(c) & (d)
\end{tabular}
\end{center}
\caption{In (a) and (b), a simple chain and tree models represented both as bi-directed graphs.
In (c) and (d), our corresponding extended factor graph representations with auxiliary variables $\mathbf Z$.}
\label{fig:factor_examples}
\end{figure}

The formulation (\ref{eq:joint_z}) has direct implications to the
simplification of the derivative-sum-product algorithm. We can now
cast (\ref{eq:pre_joint_z}) as the marginalization of
(\ref{eq:joint_z}) with respect to $\mathbf Z$, and {\it use standard
message-passing algorithms}. The independence structure now follows
the semantics of an undirected Markov network \cite{cowell:99} rather
than the bi-directed graphical model of \cite{drton:03, drton:08}. In
Figure \ref{fig:factor_examples} we show some examples using both representations,
where the Markov network independence model is represented as a factor graph.
The likelihood function can then be computed by this formulation of
the problem using black-box message passing software for junction
trees.

Now that we have the tools to compute the likelihood function,
Bayesian inference can be carried. Assume we
have for each $\phi_j(\cdot)$ a set of parameters $\{\theta_j, \mathbf
a_j\}$, of which we want to compute the posterior distribution given
some data $\mathcal D$ using a MCMC method of choice. Notice that,
after marginalizing $\mathbf Z$ and assuming the corresponding graph
is connected, all parameters are mutually dependend in the posterior
since (\ref{eq:pre_joint_z}) does not factorize in general. This
mirrors the behaviour of MCMC algorithms for the Gaussian model of
marginal independence as described by \cite{silva:09}. Unlike the
Gaussian model, there are no hard constraints on the parameters across
different factors. Unlike the Gaussian model, however, factorizations
with high treewidth cannot be tractably treated.

\section{Auxiliary Variable Approaches for Bayesian Inference}
\label{sec:auxiliary-variable}

For problems with intractable likelihoods, one possibility is to
represent it as the marginal of a latent variable model, and then
sample jointly latent variables and the parameters of interest.  Such
auxiliary variables may in some contexts help with the mixing of MCMC
algorithms, although we do not expect this to happen in our context,
where conditional distributions will prove to be quite complex.  In
\cite{silva:09}, we showed that even for small dimensional Gaussian models, the
introduction of latent variables makes mixing much worse. It may nevertheless
be an idea that helps to reduce the complexity of the likelihood calculation
up to a practical point.

One straightforward exploration of the auxiliary variable approach is
given by (\ref{eq:joint_z}): just include in our procedure the
sampling of the discrete latent vector $\mathbf Z^{(d)}$ for each data
point $d$. The data-augmented likelihood is tractable and, moreover, a
Gibbs sampler that samples each $Z_i$ conditioned on the remaining
indicators only needs to recompute the factors where variable $U_i$ is
present. The idea is straightforward to implement, but practioners
should be warned that Gibbs sampling in discrete graphical models also
has mixing issues, sometime severely. A possibility to mitigate this problem is to
``break'' only a few of the factors by analytically summing over some,
but not all, of the auxiliary $\mathbf Z$ variables in a way that the
resulting summation is equivalent to dynamic programming in a
tractable subgraph of the original graph. Only a subset will be
sampled. This can be done in a way analogous to the classic cutset
conditioning approach for inference in Markov random fields
\cite{pearl:88}. In effect, any machinery used to sample from discrete
Markov random fields can be imported to the task of sampling
$\mathbf Z$. Since the method in Section 3 is basically the result of
marginalizing $\mathbf Z$ analytically, we describe the previous method
as a ``collapsed'' sampler, and the method where $\mathbf Z$ is sampled
as a ``discrete latent variable'' formulation of an auxiliary variable sampler.

\begin{figure}[t]
\begin{center}
\begin{tabular}{ccc}
\myvcenter{\includegraphics[scale=.30]{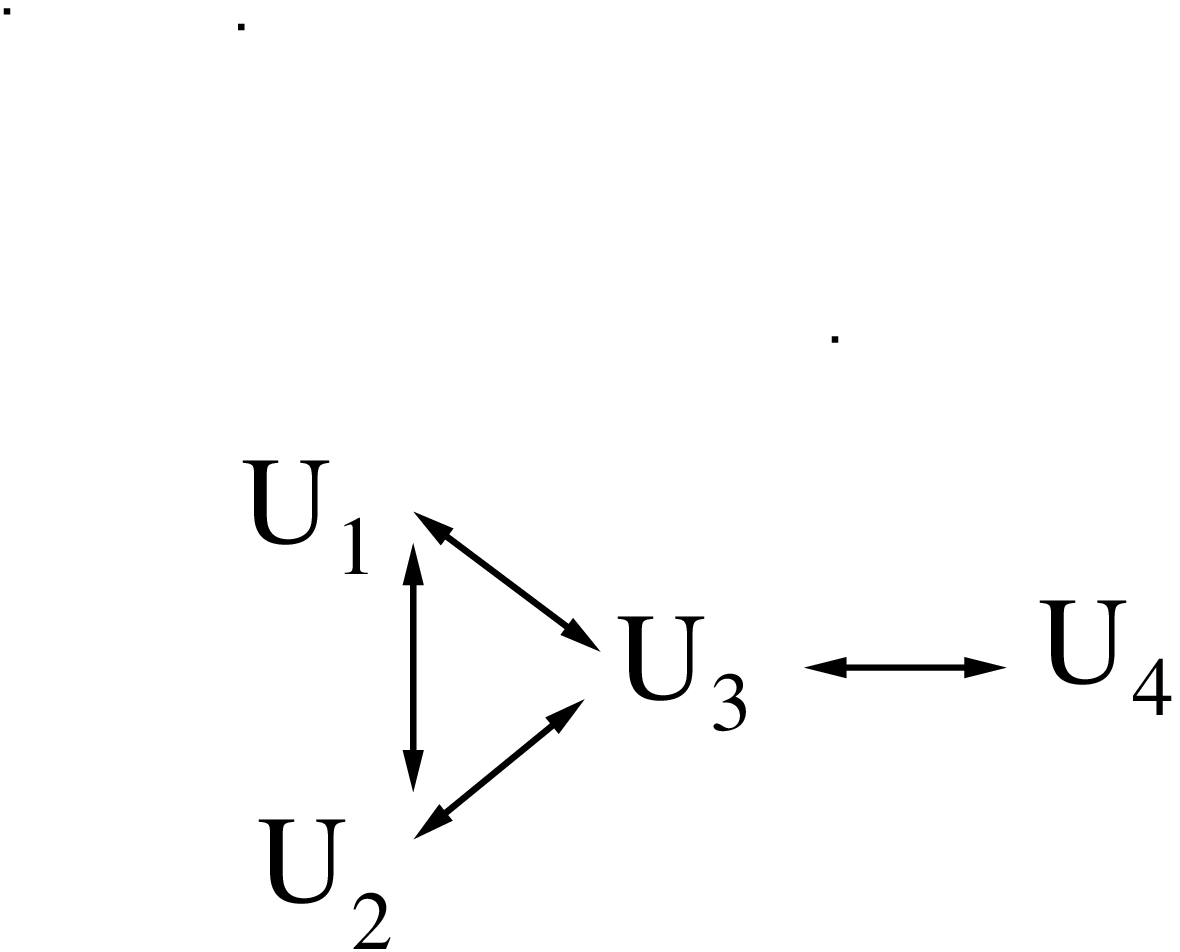}} &
\myvcenter{\includegraphics[scale=.30]{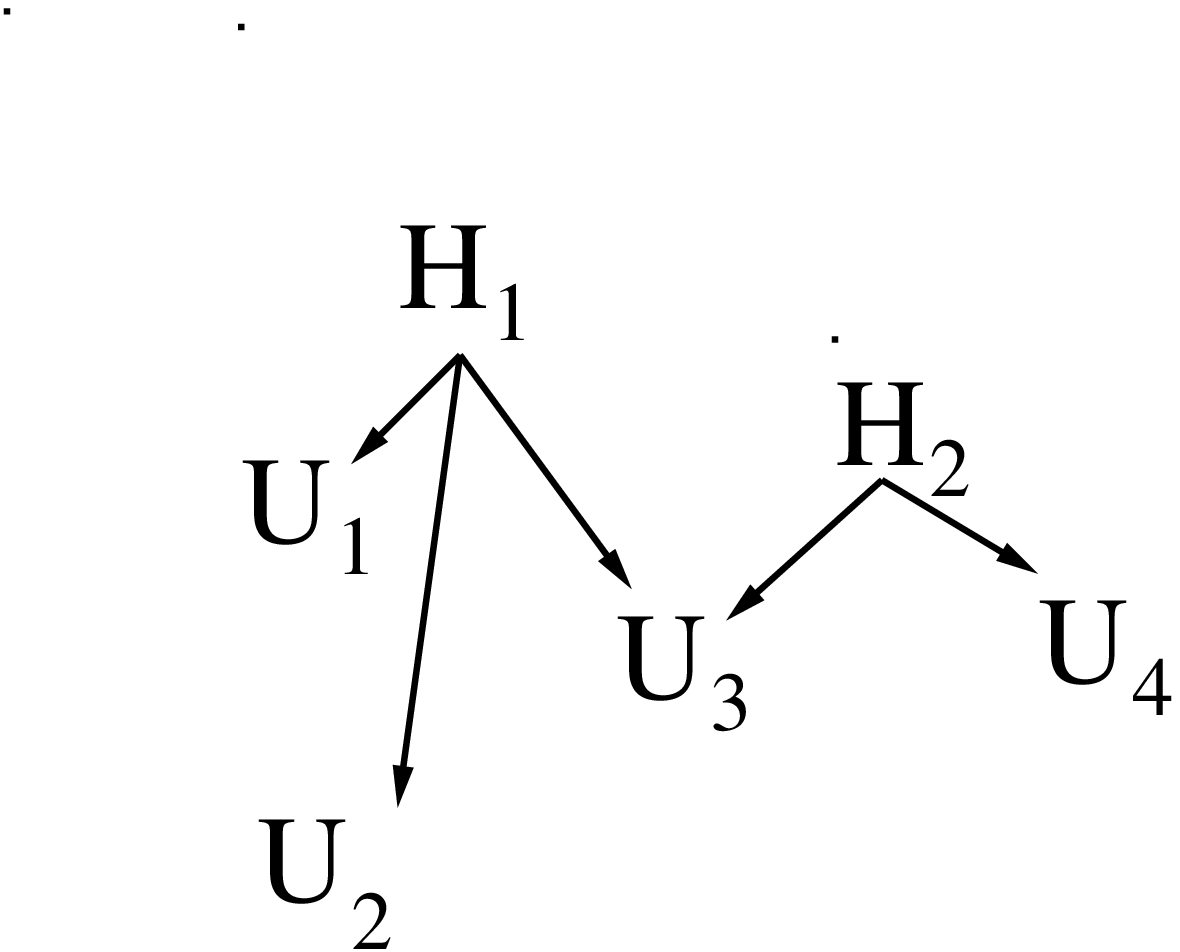}} &
\myvcenter{\includegraphics[scale=.30]{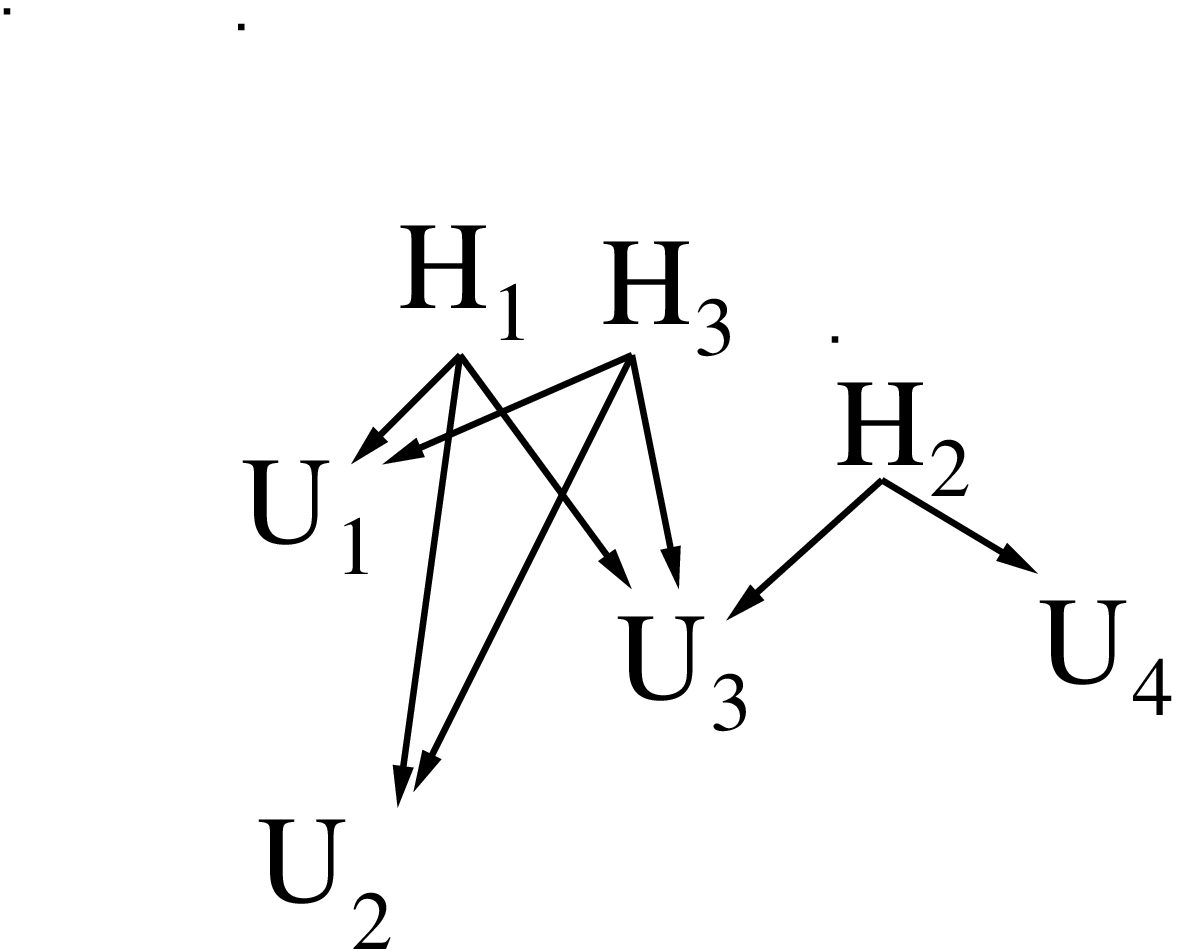}} \\
(a) & (b) & (c)\\
\end{tabular}
\end{center}
\caption{The independence constraints implied by (a) among variables $U_1$, $U_2$ and $U_3$ are also
implied by (b) and (c) according to standard graphical separation criteria (the Global Markov properties described in, e.g., \cite{richardson:02}).}
\label{fig:bidir_dag_examples}
\end{figure}

This nomenclature also helps to distinguish those two methods for yet
another third approach. This third approach is inspired by an
interpretation of the independence structure of bi-directed graph
models as given via a directed acyclic graph (DAG) model with
latent variables. In particular, consider the following DAG $\mathcal
G'$ constructed from a bi-directed graph $\mathcal G$:
i. add all variables of $\mathcal G$ as observed variables to $\mathcal G'$; ii. for each clique $S_i$ in
$\mathcal G$, add at least on hidden variable to $\mathcal G'$ and
make these variables a parent of all variables in $S_i$. If hidden
variables assigned to different cliques are independent, 
it follows that the independence constraints 
among the observed variables of $\mathcal G$ and $\mathcal G'$ \cite{richardson:02} are the same,
as defined by standard graphical separation criteria\footnote{Known as
Global Markov conditions, as described by e.g. \cite{richardson:02}.}.
See Figure \ref{fig:bidir_dag_examples} for examples.

The same idea can be carried over to CDNs. Assume for now that each CDF factor has a known representation
given by 
\[
\displaystyle
P_j(U_1 \leq u_1^{a_{1j}}, \dots, U_p \leq u_p^{a_{pj}}) = \int \left\{\prod_{i = 1}^p P_{ij}(U_i \leq u_i^{a_{ij}}\ |\ \mathbf h_j)\right\}p_{\mathbf h_j}(\mathbf h_j)\ d\mathbf h_j
\]
\noindent and that $P_{ij}$ is not included in the product if $U_i$ is not in factor $j$. Assume further that the joint distribution of $\mathbf H \equiv \cup_j \mathbf H_j$ factorizes as
\[
\displaystyle p_{\mathbf H}(\mathbf h) \equiv \prod_{j = 1}^K p_{\mathbf h_j}(\mathbf h_j)
\]

It follows that the resulting PDF implied by the product of CDFs $\{C_j(\cdot)\}$ will have 
a distribution Markov with respect to a (latent) DAG model over $\{\mathbf U, \mathbf H\}$, since
\begin{equation}
\begin{array}{rcl}
\displaystyle
\frac{\partial^p P(\mathbf U \leq \mathbf u\ |\ \mathbf h)p_{\mathbf H}(\mathbf h)}{\partial u_1 \dots \partial u_p} & = &
\displaystyle
p_{\mathbf H}(\mathbf h)\prod_{i = 1}^p\frac{\partial \{\prod_{j \in Par(i)} P_{ij}(U_i \leq u_i^{a_{ij}}\ |\ \mathbf h_j)\}}{\partial u_i}\\
&\equiv& \displaystyle
p_{\mathbf H}(\mathbf h)\prod_{i = 1}^p p_i(u_i\ |\ \mathbf h_{Par(i)})\\
\end{array}
\end{equation}
\noindent where $Par(i)$ are the ``parents'' of $U_i$: the subset of $\{1, 2, ..., K\}$ 
corresponding to the factors where $U_i$ appears. The interpretation
of $p_i(\cdot)$ as a density function follows from the fact that again
$\prod_{j \in Par(i)} P_{ij}(U_i \leq u_i^{a_{ij}}\ |\ \mathbf h_j)$
is a product of CDFs and, hence, a CDF itself.

MCMC inference can then be carried out over the joint parameter and
$\mathbf H$ space. Notice that even if all latent variables are
marginally independent, conditioning on $\mathbf U$ will create
dependencies\footnote{As a matter of fact, with one latent variable
per factor, the resulting structure is a Markov network where the edge
$H_{j_1} - H_{j_2}$ appears only if factors $j_1$ and $j_2$ have at
least one common argument.}, and as such mixing can also be
problematic. However, particularly for dense problems where the number of
factors is considerably smaller than the number of variables, sampling
in the $\mathbf H$ space can potentially sound more attractive than
sampling in the alternative $\mathbf Z$ space.

One important special case are products of Archimedean copulas. An
Archimedean copula can be interpreted as the marginal of a latent
variable model with a single latent variable, and exchangeable over
the observations. A detailed account of Archimedean copulas is given by textbooks such as 
\cite{joe:97,nelsen:07}, and their relation to exchangeable latent variable models
in \cite{marshall:88,hofert:08}. Here we provide as an example a latent variable
description of the Clayton copula, a popular copula in domains such as
finance for allowing stronger dependencies at the lower quantiles of
the sample space compared to the overall space.\vspace{0.1in}

\noindent{\bf Example 3} A set of random variables $\{U_1, \dots,
U_p\}$ follows a Clayton distribution with a scalar parameter
$\theta$ when sampled according to the following generative model
\cite{marshall:88,hofert:08}:
\begin{enumerate}
\item Sample random variable $H$ from a Gamma $(1 / \theta, 1)$ distribution
\item Sample $p$ iid variables $\{X_1, \dots, X_p\}$ from an uniform $(0, 1)$
\item Set $U_i = (1 - \log(X_i) / H)^{-1 / \theta}$ $\blacksquare$
\end{enumerate}
\vspace{0.1in}

This implies that, by using Clayton factors $C_j(\cdot)$, each associated with
respective parameter $\theta_j$ and (single) gamma-distributed latent variable $H_j$,
we obtain
\[
P_{ij}(U_i \leq u_i^{a_{ij}}\ |\ h_j) = \exp(-h_j( u_i^{-\theta_j a_{ij}} - 1))
\]
By multiplying over all parents of $U_i$ and differentiating with respect to $u_i$, we get:
\begin{equation}
\label{eq:clayton_cond_pdf}
\displaystyle
p_i(u_i\ |\ \mathbf h_{Par(i)}) = 
\left[\prod_{j \in Par(i)} \exp(-h_j( u_i^{-\theta_j a_{ij}} - 1))\right]
\left[\sum_{j \in Par(i)} \theta_j a_{ij} h_j u_i^{-\theta_j a_{ij} - 1}\right]
\end{equation}

A MCMC method can then be used to sample jointly $\{\{a_{ij}\}, \{\theta_j\}, \{\mathbf H^{(1)}, \dots, \mathbf H^{(d)}\}\}$
given observed data with a sample size of $d$. We do not consider estimating the shape of
the factorization (i.e., the respective graphical model structure learning task) as
done in \cite{silva:13}.

\section{Illustration}
\label{sec:experiments}

We discuss two examples to show the possibilities and
difficulties of performing MCMC inference in dense and sparse
cumulative distribution fields. For simplicity we treat the
exponentiation parameters $a_{ij}$ as constants by setting them to be
uniform for each variable (i.e., if $U_i$ appears in $k$ factors,
$a_{ij} = 1/k$ for all of the corresponding factors).  Also, we 
treat marginal parameters as known in this Bayesian inference exercise
by first fitting them separately and using the estimates to generate
uniform $(0, 1)$ variables.

The first one is a simple example in financial time series, where we
have 5 years of daily data for 46 stocks from the S\&P500 index, a
total of 1257 data points. We fit a simple first-order linear
autoregression model for each log-return $Y_{it}$ of stock $i$ at time
$t$, conditioned on all 46 stocks at time $t - 1$. Using the
least-squares estimator, we obtain the residuals and use the marginal
empirical CDF to transform the residual data into approximately
uniform $U_i$ variables.

\begin{figure}[t]
\begin{center}
\myvcenter{\includegraphics[scale=0.25]{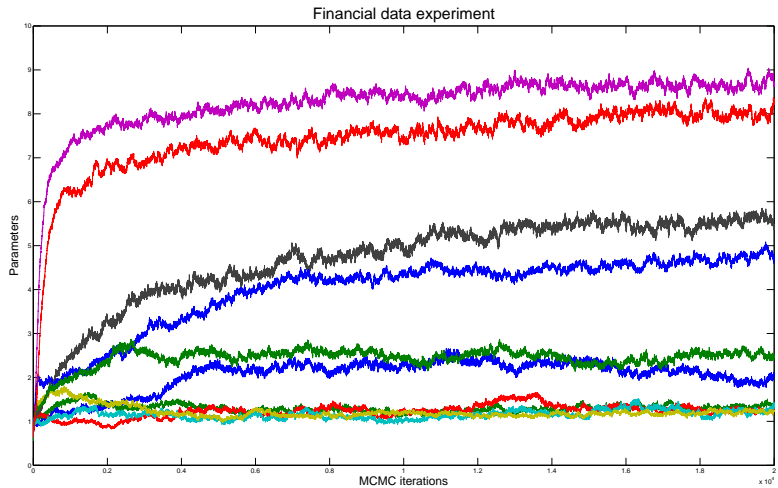}}
\end{center}
\caption{MCMC traces of the 10 parameters for the 46 log-returns data.
Convergence is slow, although each step is relatively cheap.}
\label{fig:finance}
\end{figure}

The stocks are partitioned into 4 clusters according to the main
category of business of the respective companies, with cluster sizes
varying from 6 to 15. We define a CDF field using 10 factors: one for
each cluster, and one for each pair of clusters using a Clayton copula
for each factor. This is not a sparse model\footnote{Even though it is
still very restricted, since Clayton copulas have single parameters. A
plot of the residuals strongly suggests that a t-copula would be a more
appropriate choice, but our goal here is just to illustrate the
algorithm.} in terms of independences among the observed $\{U_1,
\dots, U_{46}\}$. However, in the corresponding latent DAG model there are only 10
latent variables with each observation $U_i$ having only two parents.

We used a Metropolis-Hastings method where each $\theta_i$ is sampled
in turn conditioning on all other parameters using slice sampling
\cite{neal:03}.  Latent variables are sampled one by one using a
simple random walk proposal. A gamma $(2, 2)$ prior is assigned to
each copula parameter independently. Figure \ref{fig:finance}
illustrates the trace obtained by initializing all parameters to
1. Although each iteration is relatively cheap, convergence is substantially
slow, suggesting that latent variables and parameters have a strong
dependence in the posterior. As is, the approach does not look
particularly practical. Better proposals than random walks are
necessary, with slice sampling each latent variable being far too
expensive and not really addressing the posterior dependence between
latent variables and parameters.

Our second experiment is a simple illustration of the proposed methods for a sparse
model. Sparse models can be particularly useful to model residual
dependence structure, as in the structural equation examples of
\cite{silva:13}. Here we use synthetic data on a simple chain
$U_1 \leftrightarrow \dots \leftrightarrow U_5$ using all three
approaches: one where we collapse the latent variables and perform
MCMC moves using only the observed likelihood calculated by dynamic
programming; another where we sample the four continuous latent
variables explicitly (the ``continuous latent'' approach); and the
third, where we simply treat our differential indicators as discrete
latent variables (the ``discrete latent'' approach). Clayton copulas
with gamma $(2, 2)$ priors were again used, and exponents $a_{ij}$
were once again fixed uniformly. As before, slice sampling was used for the
parameters, but not for the continuous latent variables.

\begin{figure}[t]
\begin{center}
\begin{tabular}{c}
\myvcenter{\includegraphics[scale=.29]{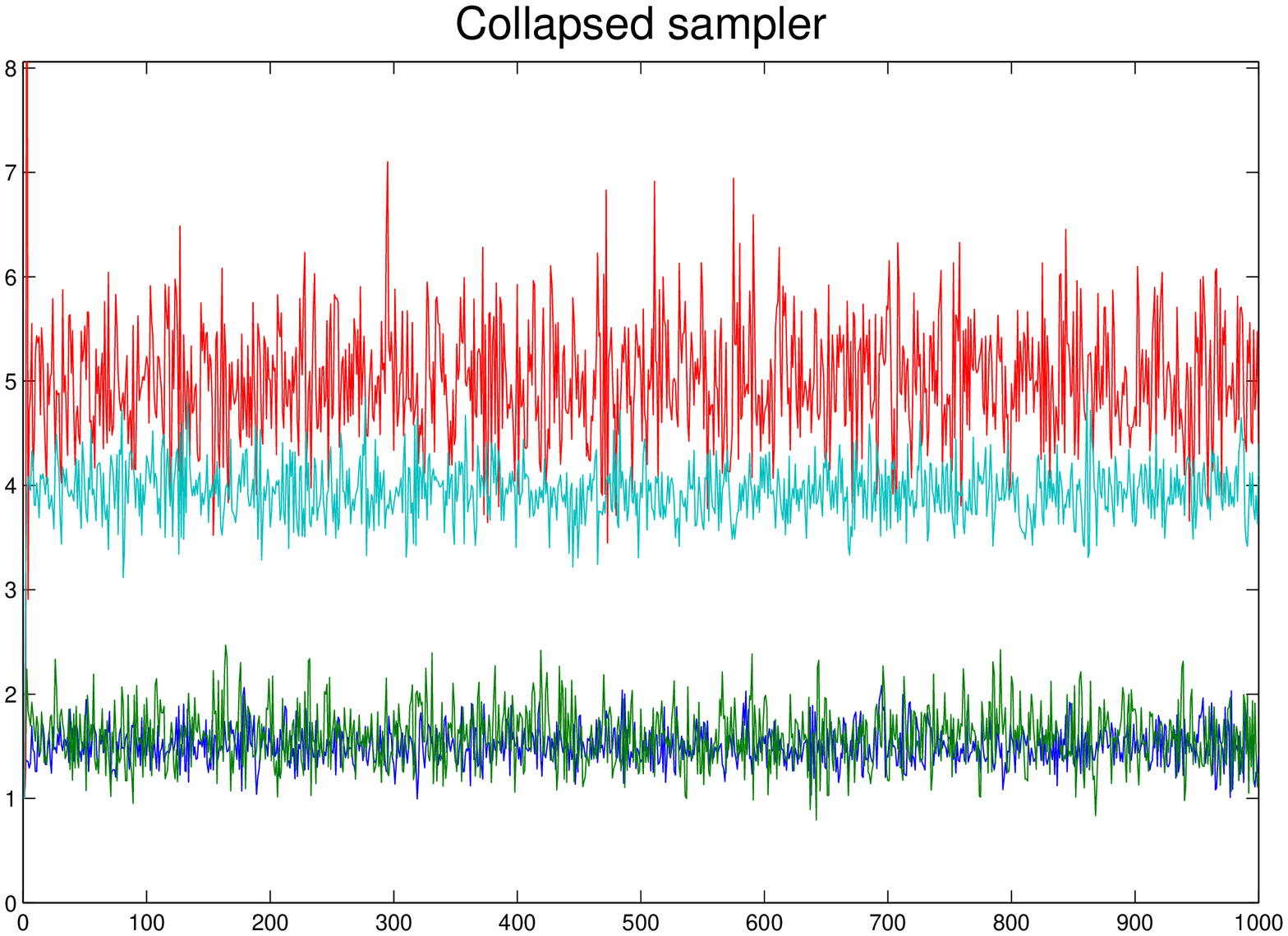}} 
\myvcenter{\includegraphics[scale=.29]{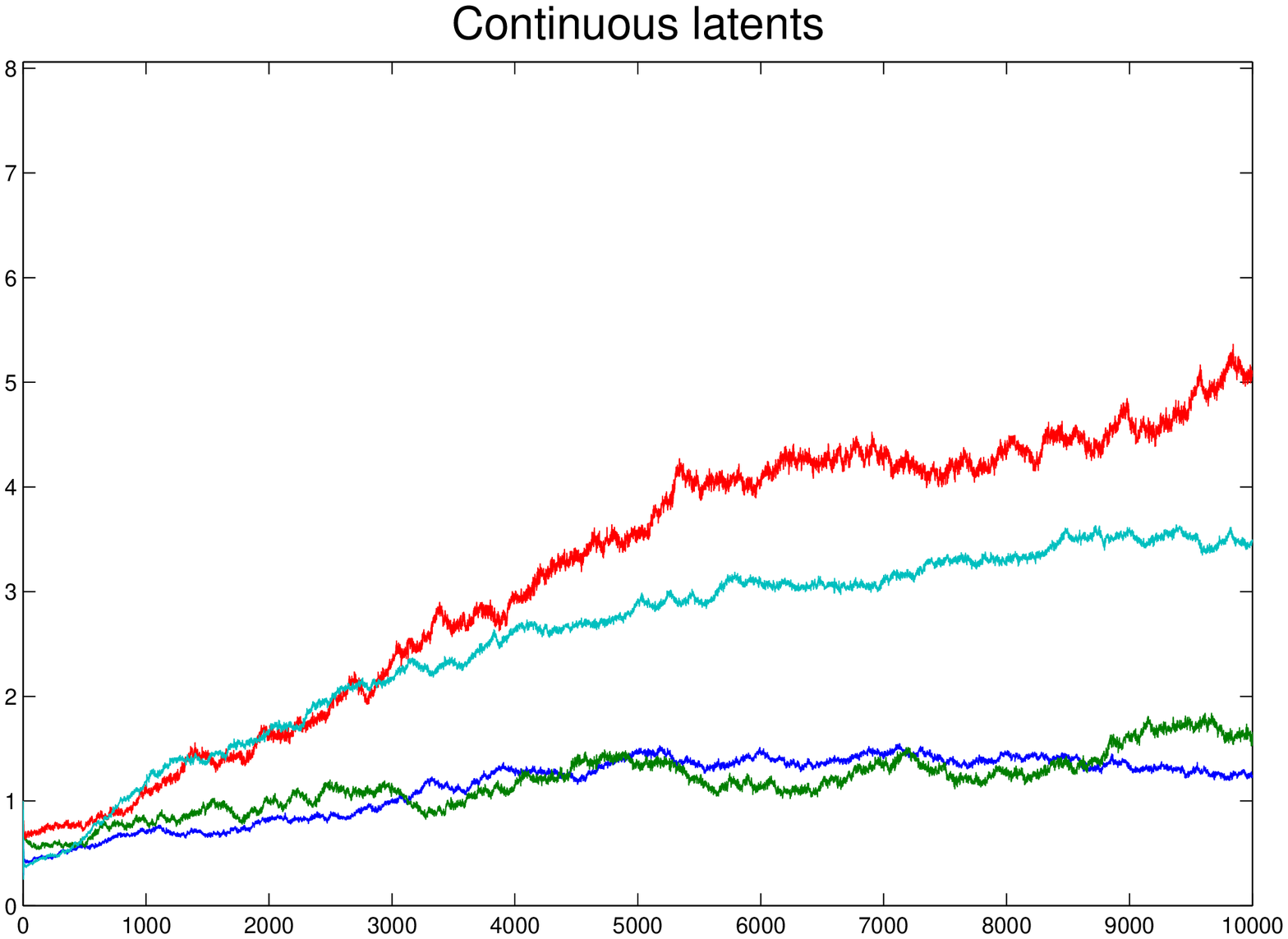}} \\
\myvcenter{\includegraphics[scale=.29]{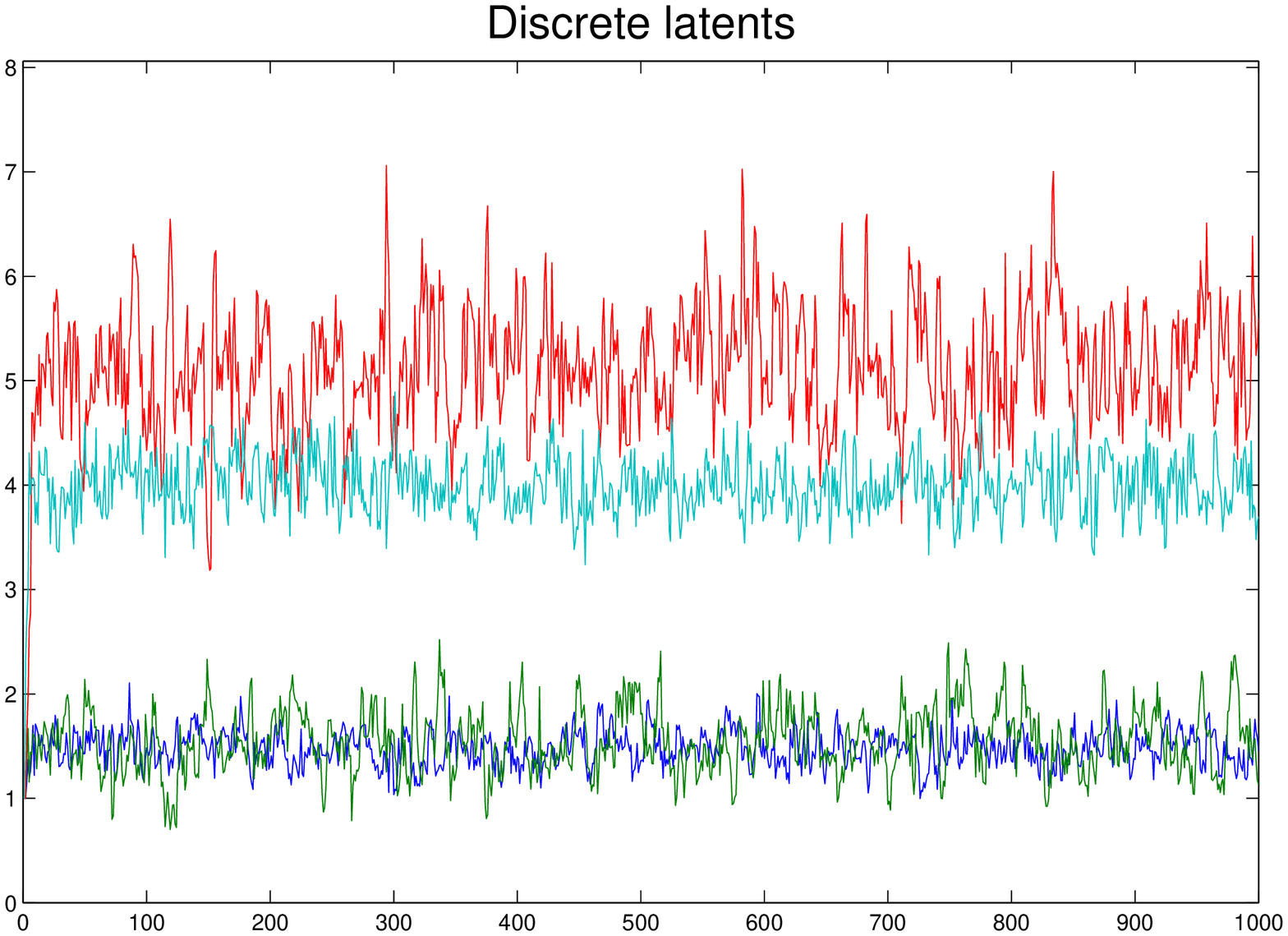}} 
\end{tabular}
\end{center}
\caption{Sampling performance for the synthetic case study using the three different methods.}
\label{fig:sampling_synth}
\end{figure}

Figure \ref{fig:sampling_synth} summarizes the result of a synthetic
study with a random choice of parameter values and a chain of five
variables (a total of 4 parameters). For the collapsed and discrete
latent methods, we ran the chain for 1000 iterations, while we ran the
continuous latent method for 10000 iterations with no sign of
convergence. The continuous latent method had a computational cost of
about three to four times less than the other two
methods. Surprisingly, the collapsed and discrete latent methods
terminated in roughly the same amount of wallclock time, but in
general we expect the collapsed sampler to be considerably more
expensive. The effective sample size for the collapsed method along
the four parameters was $(1000, 891, 1000, 903)$ and for the
discrete latent case we obtained $(243, 151, 201, 359)$.

\section{Discussion}
\label{sec:discussion}

Cumulative distribution fields provide another construction for copula
functions. They are particularly suitable for sparse models where
many marginal independences are expected, or for conditional models
(as in \cite{silva:13}) where residual association after accounting
for major factors is again sparsely located. We did not, however,
consider the problem of identifying which sparse structures should
be used, and focused instead on computing the posterior distribution 
of the parameters for a fixed structure. 

The failure of the continuous latent representation as auxiliary
variables in a MCMC sampler was unexpected. We conjecture that more
sophisticated proposals than our plain random walk proposals should
make a substantial difference. However, the main advantage of the
continuous latent representation is for problems with large factors
and a small number of factors compared to the number of variables. In
such a situation perhaps the product of CDFs formulation should not be
used anyway, and practitioners should resort to it for sparse
problems. In this case, both the collapsed and the discrete latent
representations seem to offer a considerable advantage over models
with explicit latent variable representations (at least
computationally), a result that was already observed for a similar
class of independence models in the more specific case of Gaussian
distributions \cite{silva:09}. 

An approach not explored here was the pseudo-marginal method
\cite{andrieu:09}, were an in place of the intractable likelihood
function we use a positive unbiased estimator. In principle, the
latent variable formulations allow for that. However, in a preliminary
experiment where we used the very naive uniform distribution as an
importance distribution for the discrete variables $\mathbf Z$, in a
10-dimensional chain problem with 100 data points, the method failed
spectacularly. That is, the chain hardly ever moved. Far more
sophisticated importance distributions will be necessary here.

Expectation-propagation (EP) \cite{minka:00} approaches can in
principle be developed as alternatives. A particular interesting
feature of this problem is that marginal CDFs can be read off easily,
and as such energy functions for generalized EP can be derived in
terms of actual marginals of the model.

For problems with discrete variables, the approach can be used almost
as is by introducing another set of latent variables, similarly to
what is done in probit models. In the case where dynamic programming
by itself is possible, a modification of (\ref{eq:basic-cdf}) using
differences instead of differentiation leads to a similar discrete
latent variable formulation (see the Appendix of \cite{silva:12})
without the need of any further set of latent variables. However, the
corresponding function is not a joint distribution over $\mathbf Z
\cup \mathbf U$ anymore, since differences can generate negative
numbers.

Some characterization of the representational power of products of
copulas was provided by
\cite{liebscher:08}, but more work can be done and we also conjecture
that the point of view provided by the continuous latent variable
representation described here can aid in understanding the constraints
entailed by the cumulative distribution field construction.

\section*{Acknowledgements}
The author would like to thank Robert B. Gramacy for the financial data.
This work was supported by a EPSRC grant EP/J013293/1.

\bibliographystyle{natbib}

\end{document}